\documentclass[utf8, 10pt, a4paper]{article}
\usepackage{lrec}
\usepackage{graphicx}
\usepackage{tabularx}
\usepackage{soul}
\usepackage{CJKutf8}  % added by Aaron
\usepackage{multicol, blindtext}
\usepackage[utf8]{inputenc}
\usepackage{epstopdf}
\usepackage{url}
\usepackage{comment}

\title{MultiMWE: Building a Multi-lingual Multi-Word Expression (MWE) Parallel Corpora}

\name{Lifeng Han$^1$, Gareth J.F. Jones$^1$ and Alan F. Smeaton$^2$}

\address{$^1$ ADAPT Research Centre \\
         $^2$ Insight Centre for Data Analytics \\
         School of Computing, Dublin City University, Glasnevin, Dublin 9, Ireland\\
         lifeng.han@adaptcentre.ie; \{gareth.jones, alan.smeaton\}@dcu.ie\\}

\abstract{%鴻鵠之志
Multi-word expressions (MWEs) are a hot topic in research in natural language processing (NLP), including topics such as MWE detection, MWE decomposition, and research investigating the exploitation of MWEs in other NLP fields such as Machine Translation. However, the availability of bilingual or multi-lingual MWE corpora is very limited. The only bilingual MWE corpora that we are aware of is from the PARSEME (PARSing and Multi-word Expressions) EU Project. This is a small collection of only 871 pairs of English-German MWEs. In this paper, we present multi-lingual and bilingual MWE corpora that we have extracted from root parallel corpora. Our collections are 3,159,226 and 143,042 bilingual MWE pairs for German-English and Chinese-English respectively after filtering. We examine the quality of these extracted bilingual MWEs in MT experiments. Our initial experiments applying MWEs in MT show improved translation performances on MWE terms in qualitative analysis and better general evaluation scores in quantitative analysis, on both German-English and Chinese-English language pairs. We follow a standard experimental pipeline to create our MultiMWE corpora which are available online. Researchers can use this free corpus for their own models or use them in a knowledge base as model features. \\ \newline  }

\begin{document}

\maketitleabstract

\section{Introduction}

%\footnote{Accepted to LREC2020 \url{https://lrec2020.lrec-conf.org/en/}}
%*\footnote{Accepted to LREC2020. Language Resources and Evaluation (LREC), Marseille, France.}
The use of multi-word expressions (MWEs) has  become a  hot topic in research in the field of natural language processing (NLP). Topics of interests in MWEs include  issues such as  MWE detection \cite{maldonadoHanMoreau2017detection}, MWE decomposition, and the integration of MWEs into other NLP applications such as Machine Translation (MT). However, to support research into the multilingual use of MWEs, the availability of bilingual or multi-lingual MWE corpora is very limited. The only freely available bilingual MWE corpora that we are aware of, at the submission time, is from the PARSEME (PARSing and Multi-word Expressions) EU Project\footnote{https://typo.uni-konstanz.de/parseme/}. This corpus is quite small, containing only 871 pairs of English-German MWEs.
In this paper we present details of multi-lingual and bilingual MWE corpora that we have extracted from  parallel corpora. We examine the quality of these extracted bilingual MWEs in MT experiments. Results for our initial experiments of applying MWEs into the MT process show improved translation evaluation scores on German-English and Chinese-English language pairs. These initial results justify further development of MWEs and their use in MT and potentially other NLP applications. We follow a standard experimental  pipeline \cite{rikters2017mwe} to extract our bilingual MWEs. Our MultiMWE corpora are freely available online\footnote{https://github.com/poethan/MWE4MT}.

This paper is organized as follows: Section 2 provides some background knowledge on MWEs and MT, Section 3 lists some related works, Section 4 is the MultiMWE corpora extraction procedure, Section 5 presents some experiments on MWE integration into MT, and Section 6 is our discussion and conclusions.

\section{Background}
In this section, we introduce the concept of MT and MWE, and illustrate their connection with examples. This provides background to the motivation for the development of the MultiMWE corpora which forms the subject of this paper.

\subsection{Machine Translation and Multiword Expressions}
\begin{CJK*}{UTF8}{bsmi}

MT methods seek to translate one  human language into another one. MT belongs to a branch of computational linguistics (CL) and artificial intelligence (AI), in which researchers try to use computational modeling to address linguistic text translation problems. It is a very challenging task for MT to achieve both accuracy of translated information and fluency at the level of a human expert's performance or what  linguists expect as output. There are many reasons for this, one of which is that the use of MWEs presents a significant obstacle for a machine to learn and generate human languages in a natural form. We use three examples to illustrate the importance of correct use of MWEs in MT.
We use ZH/Zh to represent Chinese, and EN/En as English. We use pinyin (pīnyīn) to annotate the Báihuà Chinese for its pronunciation and tones (phoneticism). The MT outputs in the examples were from Google Translator engine \cite{google2017attention}, which represented one of state-of-the-art neural models\footnote{tested on 2018/10/26}. 
%We illustrate this with three examples with MWE issues as a reflection of how this is important in MT.
\subsubsection{Example-I: Báihuà\, (白話)}
As a first example, we show how important it is to understand  Chinese expression patterns in order to express the correct tense and overall information in a sentence.

%\begin{CJK*}{UTF8}{bsmi}

\begin{figure}[!t]
\centering
\includegraphics*[height=1.0in]{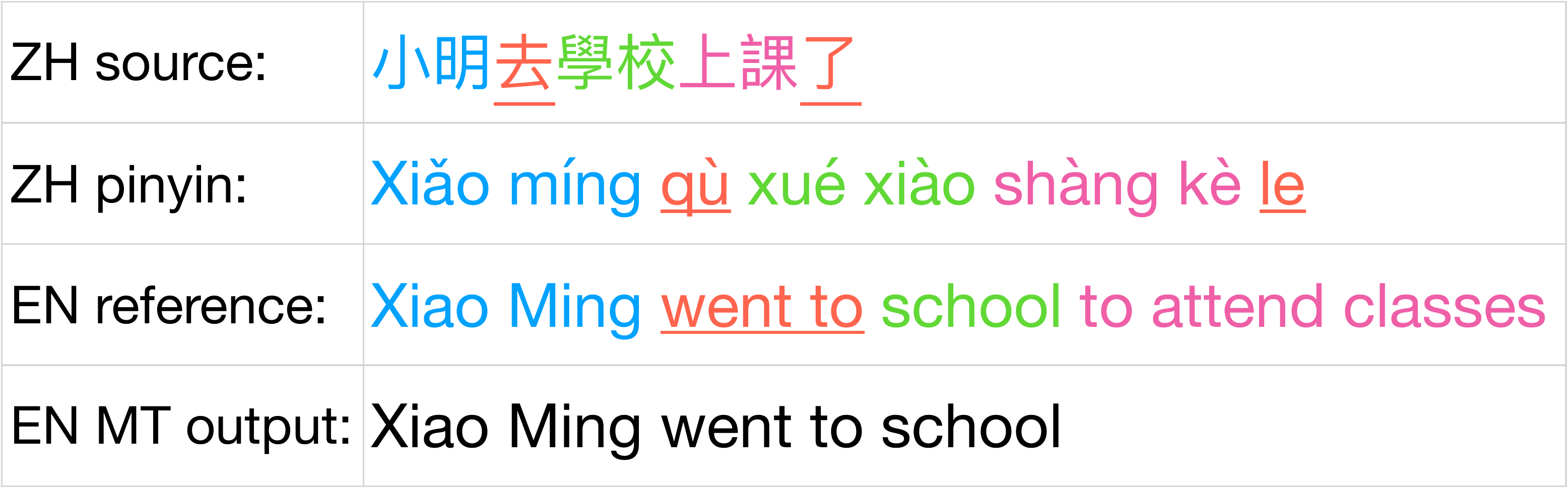}
%[height=3.1in,width=2.6in]
\caption{Example-I Translation of Chinese (Báihuà, modern Chinese) to English. The pattern ``去\, (qù) ... 了\, (le)" is a dis-continuous MWE often used to express the past tense ``went to do something". Here it is used to express ``went to somewhere (school) for something (attending classes)".}
\label{fig:example_MT-I}
\end{figure}

Let us examine a simple plain example ``小明去學校上課了 \,(phoneticism: xiǎo míng qù xué xiào shàng kè le)" of modern Chinese `白話\, (Báihuà)', compared with ancient `文言\, (Wényán)' of which we will show one example later, to English MT as in Figure \ref{fig:example_MT-I}.  In this simple example, the MT output has lost the \textit{aim} of Xiao Ming's action to go to school, i.e., what is his purpose to go there (\textit{to attend  classes}). This reflects an overall loss of adequacy. In Chinese, there is no direct past tense in the verb, so the MT needs to acquire the knowledge of language expression patterns to be able to express the tense information and purpose of the action here. The Chinese pattern ``去\, (qù) ... 了\, (le)" is a simple \textbf{dis-continuous Chinese MWE} used to express a past tense action (went to do something, went to somewhere). 
\end{CJK*}

\begin{CJK*}{UTF8}{bsmi}

\end{CJK*}
\begin{CJK*}{UTF8}{bsmi}
\subsubsection{Example-II: Poem\, (詩歌)}
%\begin{CJK*}{UTF8}{bsmi}
For the second example, we will see how correct understanding of Chinese MWEs can assist disambiguation in machine learning. Conversely, the failure to understand these MWEs can lead to an incorrect  translation of the ambiguous Chinese character even in  very well aligned poem sentences.

\begin{figure}[!t]
\centering
\includegraphics*[height=1.4in]{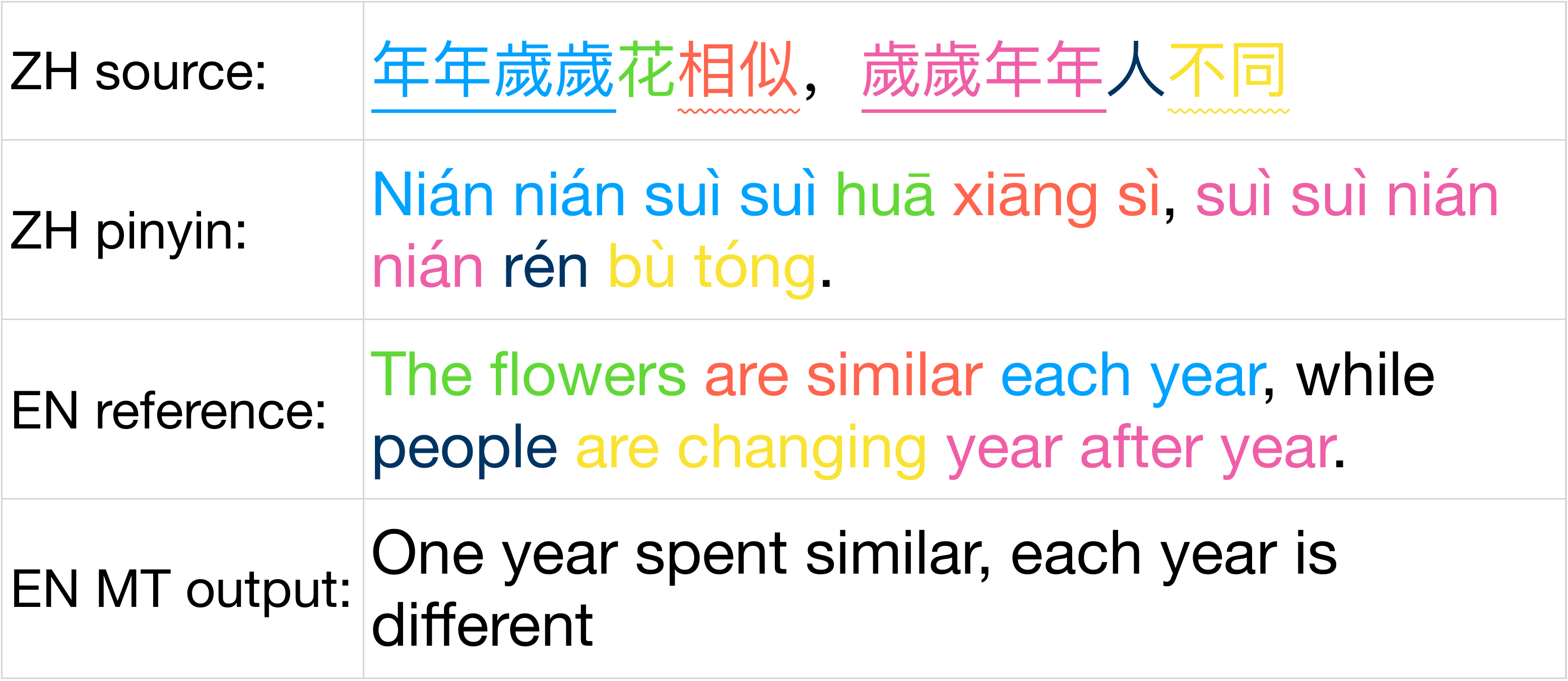}
%[height=3.1in,width=2.6in]
\caption{Example-II Translation of Chinese (poem) to English. The terms ``年年歲歲\, (nián nián suì suì)" and ``歲歲年年 \,(suì suì nián nián)" are continuous MWEs. ``相似\, (xiāng sì)" and ``不同\, (bù tóng)" are words with clear boundaries.}
\label{fig:example_MT-II}
\end{figure}

The second example sentence is ``年年歲歲花相似，歲歲年年人不同 \, (phoneticism: nián nián suì suì huā xiāng sì, suì suì nián nián rén bù tóng)" from one poem ``《代悲白頭翁》" of Tang Dynasty by Xiyi Liu\footnote{劉希夷\, in Chinese, 651—679, who died at early age due to this famous poem he wrote. https://zh.wikipedia.org/wiki/劉希夷}, shown in Figure \ref{fig:example_MT-II}. The source Chinese sentence is a kind of popular saying in a poetic and rhythm format with metaphor. In this example, ``年年歲歲" and ``歲歲年年" are aligned continuous MWEs saying `each year'; ``花" and ``人" are aligned as subject `flower' and `human'; ``相似" and ``不同" are aligned as action/status `(Flowers) stay the same' while `(Humans) are changing'. For the first half of the sentence, the MT engine translated `年年歲歲\, (nián nián suì suì)' into `one year', `花\, (huā)' into `spent', and `相似\, (xiāng sì)' into `similar'. While the translation of the MWE `年年歲歲\, (nián nián suì suì)' is a credible attempt since it should be `each year', the translation of `花\, (huā)' is totally wrong in this sentence since it refers to `flower'. This is due to the ambiguity problem in language, since the Chinese character `花\, (huā)' also carries a meaning of `spend' in other situations such as in this example of Chinese Báihuà, `我花一百，你呢？\, (Wǒ huā yībǎi, nǐ ne?)' means `I spend one hundred, how about you?'. In the second half of the MT translation, `each year is different' loses the translation of the character `人\, (rén, meaning people)', i.e. \textit{people} are different each year. This reflects an overall \textbf{loss of adequacy} which is similar to the situation of example one.

% \url{https://zh.wikipedia.org/wiki/???}

%ZH source sentence: ??????????????? 

%ZH source pinyin: Nián nián suì suì hu? xi?ng sì, suì suì nián nián rén bù tóng

%EN reference sentence: The flowers are similar each year, while people are changing year after year.

%EN MT output: One year spent similar, each year is different. (by Google Translator, 20181026th)

In this example, if the MT model can understand the MWEs well, i.e., ``年年歲歲" aligned to ``歲歲年年" and ``相似" aligned to ``不同", then it is easier to acquire the knowledge that ``花" is aligned to ``人". Since ``人" is a subject here meaning ``person/people", ``花" should also be a noun or pronoun, which will tell the machine to translate it with higher probability into ``flower (noun)" instead of ``spent (verb)". 
We assume that the correct recognition and translation of surrounding MWEs, in general, can help the MT model to understand the  sentences better overall and improve the translation of ambiguous Chinese characters.

\end{CJK*}
\begin{CJK*}{UTF8}{bsmi}
\subsubsection{Example-III: Wényán\, (文言)}
%\begin{CJK*}{UTF8}{bsmi}
As a third example, similar to example one (Chinese Báihuà), we  show how the MT model fails to translate an ancient Chinese Wényán sentence due to the lack of \textbf{Chinese pattern expression} knowledge. Even though it is still a popular saying, the translation of this Wényán sentence is much worse than the translation of current Chinese Báihuà. This example also contains the \textit{multi-character named entity} information as one kind of MWE.

\begin{figure}[!t]
\centering
\includegraphics*[height=1.6in]{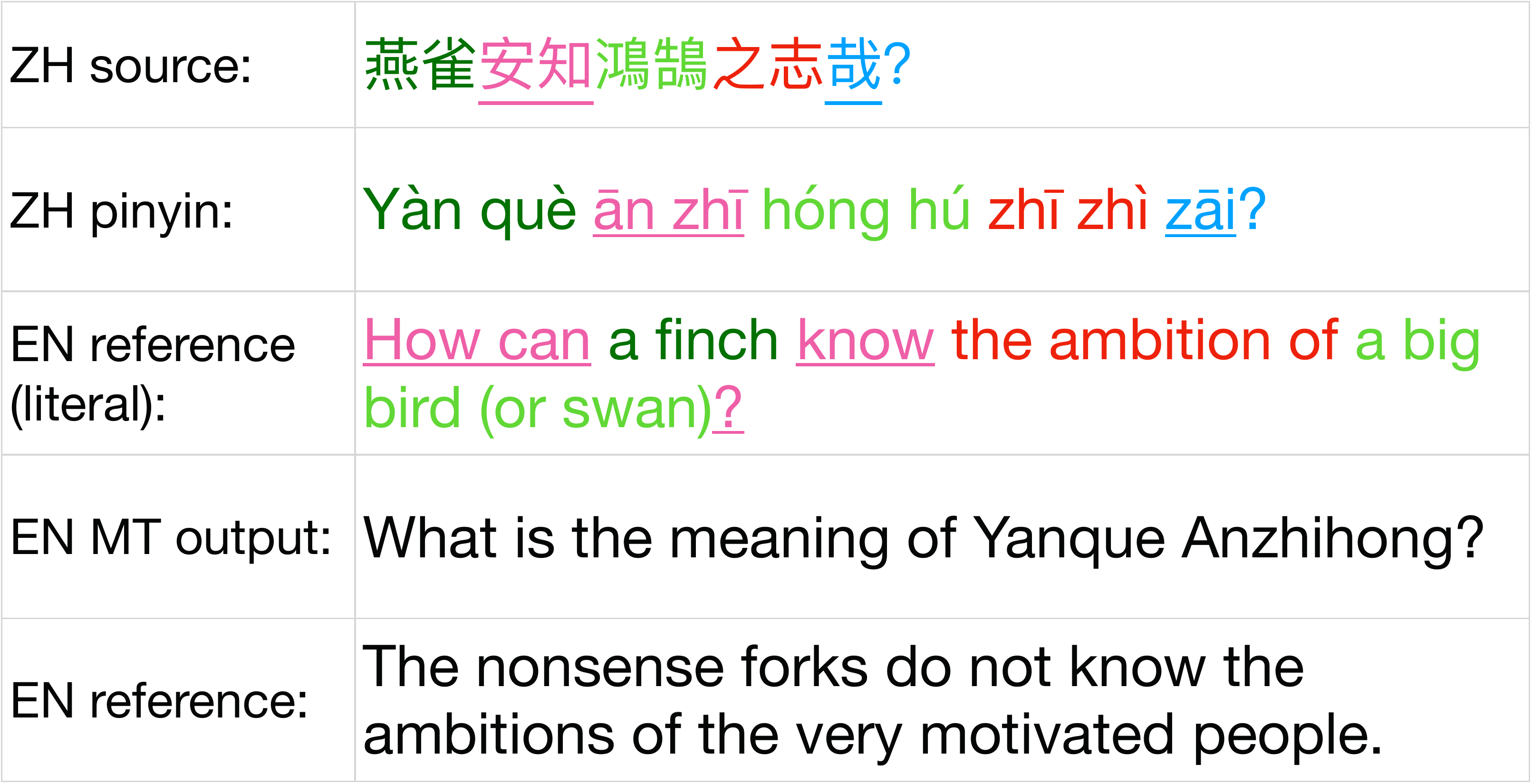}
%[height=3.1in,width=2.6in]
\caption{Example-III Translation of Chinese (Wényán) to English. ``A 安知\, (ān zhī) something 哉\, (zāi)" is a pattern to express ``how can A know something" or ``A does not know something". ``燕雀\, (yàn què)" and ``鴻鵠\, (hóng hú)" are named entities as one popular kind of MWE, and ``之志\, (zhī zhì)" is a fixed pattern expressing ``\textit{someone}'s ambition".}
\label{fig:example_MT-III}
\end{figure}

The third example, shown in Figure~\ref{fig:example_MT-III}, is a translation of the ancient Chinese 文言\, (Wényán) idiom/metaphor expression to English: ``燕雀安知鴻鵠之志哉？\, (phoneticism: yàn què ān zhī hóng hú zhī zhì zāi?)" from the book ``《史記》" \footnote{From Han Dynasty, 206 BC–220 AD \url{https://en.wikipedia.org/wiki/Records_of_the_Grand_Historian}} by Sima Qian. This Chinese expression is often used in modern language to express someone's feelings in both verbal and written format. The MWE pattern ``A 安知\, B 哉？" is used to express ``how can A know B?" or ``A does not know B". This metaphor is used to describe that some not serious or very common folks do not know the ambition or great plan of other very motivated ones.

The MT output is  poor due to the model  not understanding the meaning of the entities ``燕雀\, (yàn què, meaning finch)" and ``鴻鵠\, (hóng hú, meaning big bird, swan)", the \textbf{fixed/patterned expressions} ``安知\, (ān zhī, meaning `how to know' or `do not know')"  and ``之志\, (zhī zhì, meaning \textit{someone}'s ambition)". In the MT output, we can see that it keeps `Yanque' in the form of the original  Chinese pinyin pronunciation. This may be due to the MT system  not acquiring this meaning equivalent word from its training data. The MT output also failed by  putting `Anzhihong' together the pinyin pronunciation of the three Chinese characters `安知鴻', which makes no sense at all, since `安知\, (ān zhī)' is one term (patterned expression) and `鴻\, (hóng)' should be part of another term (named entity) `鴻鵠\, (hóng hú)'. The failure to correctly interpret these kinds of expressions  presents an obstacle to effective MT.

\end{CJK*}

\subsection{Multiword Expressions}

Researchers in computational linguistics have defined MWEs in multiple ways. However, in general, these definitions agree on the following: 
\textit{a MWE shall be a term including  several words to express a specific concept, which is able to be decomposed, and the words combined together as an MWE are syntactically, semantically or pragmatically (some people may add `statistically' from the computational view) idiosyncratic in nature.} \cite{Sag2002MWE,Tim2010mwe,huning2015MWEs}

The categories of MWEs can include idioms, compound nouns, or word combinations from different kinds of Part-of-Speech (PoS) such as verb-particle or proper names.  MWEs can be classified into lexical phrases and institutionalized phrases \cite{Sag2002MWE}. Lexical phrases include fixed or semi-fixed expressions and syntactically-flexible expressions. For instance, for verb-particle structures, there are examples: pick up, give up, put on, take off, take over, etc. For idioms, there are `you are the apple of my eye', `kick the bucket', etc. (just to list a few). MWEs can be \textbf{continuous} or \textbf{dis-continuous} in presentation.  Continuous MWEs are words grouped together without gaps, while  discontinuous ones have gaps in the overall expression, e.g. some common words inserted into  MWE word groups, for instance, pick \textit{someone} up.

% ??? ???? ?·??????????????????????????????????? http://chengyu.t086.com/cy9/9062.html ????????????????????????shù dà zh?o f?ng ???	???? ???	???? ????

%compound nouns. toothpaste, blackboard, monthly ticket, swimming pool, underground, dry-cleaning, output, full moon, overweight, outgoing, somewhere https://www.learnenglish.de/grammar/nouncompound.html  [https://www.learnenglish.de/basics/compoundwords.html]

%For instances, for compound nouns, they can have different kinds of combinations: toothpaste (noun+noun), full moon (adjective+noun), swimming pool (verb+noun), monthly ticket (adverb+noun), output (preposition+verb), haircut (noun+verb), etc.

%\begin{CJK*}{UTF8}{bsmi}

%For idioms, let's see some Chinese examples: ???? (shù dà zh?o f?ng). ? means tree/trees, ? means big/large, ? means bring in/invite, and ? means wind. The literal translation can be ``big trees bring in wind." The meaning of this metaphor is to describe that when people become famous or rich, it usually easily got attention from others therefore easier to bring troubles. So the full understanding of this idiom and MWE will be very helpful to understand the whole meaning of the sentence or documents. On the other hand, if the model can not recognize this whole MWE and splitted it into different words, it is more possible to lead to misunderstanding of the whole sentence.

%\end{CJK*}

% ``lexicalized phrases (fixed, semi-fixed and syntactically flexible expressions) and institutionalized phrases."

\begin{CJK*}{UTF8}{bsmi}

For noun phrase MWEs, we can find in example 2, ``年年歲歲\, (nián nián suì suì, noun noun noun noun)" and ``歲歲年年\, (suì suì nián nián, noun noun noun noun)" meaning `each year' or `every year'. All  four Chinese characters are individually nouns, but together they can form a phrase that can be used in an adverbial function in the sentence. We can also find in example 3, ``燕雀\, (yàn què, noun noun)" and ``鴻鵠\, (hóng hú, adjective+noun)" which are noun phrases meaning finch and swan and they are also named entities.

% for nianniansuisui - ?????????????????????\textit{??}??????????????? ? ... 

%ZH source sentence: ??????????????? 
%ZH source pinyin: Nián nián suì suì hu? xi?ng sì, suì suì nián nián rén bù tóng

% ZH (Wényán) source sentence: ??????????
%ZH source pinyin: Yàn què ?n zh? hóng hú zh? zhì z?i? 
%EN (literal translation): How can a finch know the ambition of a big bird (or swan)?
For verbal phrase MWEs in Chinese, we can also find ``安知\, (ān zhī)" from example 3, meaning `how to know / do not know'. For fixed-expressions, we can find ``之志\, (zhī zhì, particle+noun)" meaning ``(someone)'s goal/ambition" as a noun phrase MWE.

Examples 1, 2 and 3 illustrate that it can make computational models much easier to correctly interpret the whole sentence if they can recognize continuous and discontinuous MWEs first or during model learning.
\end{CJK*}

One recent book about MWE, MT and combined research  including rule-based, example-based, statistical and neural MT is \cite{mitkov2018MWUMTTT}. This introduces MWE research focused on various kinds of MWE types and covering different languages, including English-Basque (noun+verb), French-Romanian (verb+noun collocation), named entities (Persian, Turkish, Arabic, Pashto), German nominal compounds, Dutch compound splitting, and Croatian idioms.
% is reviewed by \cite{Haque2019MWUMTTT},

\subsection{MWEs in MT}
MWEs play a significant role in  language understanding and processing tasks, including MT. This is due to their very frequent appearances and their concept specific presentation. How to recognize MWEs correctly and translate them in a meaning-preserving way, instead of merely surface word translation is a challenging task. This section introduces existing research work in this area.

MWEs in MT are related to word sense \textbf{disambiguation} (WSD) \cite{wordsense2005mt,wsd4mt2007}, phrase \textbf{boundary} detection, and semantics \cite{van2007semantics}. Instead of a single word case in WSD, MWEs are multiple-word expressions, which can be translated in an awkward way if the translation model cannot translate the actual meaning of the MWE in the sentence and context, such as  metaphorical MWEs (`apple of someone's eyes', `kick the bucket', listed as simple examples). Addressing MWE translation also addresses the \textbf{semantic} aspects of translation in addition to issues of syntax, e.g. MWE boundary (detection) and its affects on overall sentence understanding. For instance,  example 3 shows how the MT model produced very poor output due to not recognizing MWE boundaries well. Investigations into  WSD have been carried out in the context of research into Neural Machine Translation (NMT).  \cite{wordsense2018nmt} shows that despite its general effectiveness NMT does not provide a full solution to the challenges of WSD. From this result, we  have an indication of how challenging it is to find a solution to the issues of  \textit{multi-word sense disambiguation} is in MT. It is highlighted in \cite{rareWordsense2017nmt} that WSD of  \textbf{rare words} is especially  difficult in NMT. The most recent work  exploring word senses in NMT e.g. with the Transformer model includes \cite{WStransformer2018nmt}.

%There are curtain amount of attentions that have been drawn to MWEs in MT since the SMT era.

\vspace{0.4cm}

\subsubsection{SMT+MWE}

We introduce research work combining SMT and MWEs here.

The earliest work that combined MT with MWEs includes \cite{lambert2005mwe}. This applied bilingual MWE pairs to modify the word alignment procedure of MT to improve translation quality on an English-Spanish corpus. The modification function on alignment was achieved by grouping the MWEs as one token before training.

% ``In (Lambert and Banchs, 2005), authors introduce a method in which a bilingual MWES lexicon was used to modify the word alignment in order to improve the translation quality. In their work, bilingual MWES were grouped as one unique token before training alignment models. They showed on a small corpus, that both alignment quality and translation accuracy were improved. However, in a further study, a lower BLEU score is re- ported after grouping MWES by part-of-speech on a large corpus (Lambert and Banchs, 2006)." // %P. Lambert and R. Banchs. 2005. Data inferred multi-word expressions for statistical machine translation. In Pro- ceedings of MT SUMMIT.
% 2006_Grouping multi-word expressions according to part-of-speech in statistical machine translation. P. Lambert and R. Banchs. 2006. Grouping multi-word expressions according to part-of-speech in statistical ma- chine translation. In Proceedings of the Workshop on Multi-word Expressions in a multilingual context. 

Further work includes \cite{Ren2009mwe} which integrated bilingual Chinese-English MWEs into the SMT toolkit Moses, \cite{Bouamor2012IdentifyingBM} which designed models to extract continuous MWEs and integrated them into the Moses system for French-English translation, and \cite{Skadina2016MultiwordEI} which discussed various MWEs in English-Latvian MT. Recent interesting work \cite{EBRAHIM2017111mwe} focused on phrasal verb MWEs  in Arabic-English phrase-based SMT. Similar to the work above, we use different bilingual MWE extraction workflows and integrate the extracted MWE pairs into training corpora. %BLEU score showed that the baseline actually achieved the best score without  integration of MWEs. 
% the BLEU score showed that the baseline system outperform all the other experiments, we believe that with a linguistic modification in the alignment step the MWE detection and integration technique we used would have higher score.

%\cite{Ren2009mwe} integrated bilingual Chinese-English MWEs into the SMT toolkit Moses systems.

%Bouamor et al. (2011)%% D. Bouamor, N. Semmar, and P. Zweigenbaum. 2011. Improved statistical machine translation using multi-word expressions. In Proceedings of MT-LIHMT, Barcelona, Spain.

% 2012-Identifying bilingual Multi-Word Expressions for Statistical Machine ``we describe a strategy for detecting translation pairs of MWEs in a French-English parallel corpus. In addition we introduce three methods aiming to integrate extracted bilingual MWES in MOSES, a phrase based Statistical Machine Translation (SMT) system"

%%to cite %2016 Multi-word Expressions in English-Latvian Machine Translation 

\vspace{0.4cm}

\subsubsection{NMT+MWE}

This section introduces work on the incorporation  of MWEs in NMT.

%Some MT researchers have explored the use of   \textbf{word composition} knowledge in their models, especially for  Western languages. For instance, \cite{matthews2016synthesizing} developed an MT model for English-German and English-Finnish which includes consideration of synthesising compound words.

MWEs can appear in different kinds of examples, such as \textbf{Names Entities (NE)} \cite{han2013chinese} when the entities appear as a chunk of several words. In \cite{Li2016neuralname}, the author applied a character level sequence to sequence modeling to translate  named entities and then integrated this into an overall NMT system on a Chinese-to-English task. This model was originally designed to solve the unseen word translation issue, but the results show that NEs in NMT  helps to improve overall translation effectiveness as measured by BLEU score. It showed the model can derive higher quality named entity alignment in the training corpus.  Similarly, the work in \cite{ugawa2018neural} focuses on the difficulty of translation of compound words in the source language, by introducing an encoder for the input word at the NE tag level at each time step. Furthermore, they designed a chunk-level LSTM above the word-level one to capture the compound named entity.
% ``2016 Neural Name Translation Improves Neural Machine Translation"

%To alleviate these problems, the encoder of the proposed model encodes the input word on the basis of its NE tag at each time step, which could reduce the ambiguity of the input word. Furthermore, the encoder introduces a chunk-level LSTM layer over a word-level LSTM layer and hierarchically encodes a source-language sen- tence to capture a compound NE as a chunk on the basis of the NE tags.   ``2018-Neural Machine Translation Incorporating Named Entity. COLING"

In \cite{rikters2017mwe}, the authors showed how enhancing MWEs knowledge by adding them into a corpus can improve NMT even  with very simple integration. For example,  they extracted  bilingual MWEs in the corpus and added  bilingual MWEs pairs and  sentence pairs that included the MWEs into a parallel corpus to train the NMT system in English to Czech and English to Latvian MT. The authors developed an alignment visualization tool to view the improvement in MWE alignment. The neural network platform they used is from Neural Monkey \cite{NeuralMonkey2017}. Our corpora construction procedure follows the pipeline designed in this work. % Helcl, J. and Libovicky?, J. (2017). Neural Monkey: An open-source tool for sequence learning. The Prague Bulletin of Mathematical Linguistics, (107):5?17.

%cut: \subsection{Motivation of MT+MWE}

%Native speakers of one language often use many customised terms  from this specific language that are not easily captured by others who speak the mother tones of other languages. The root of the customised terms, such as idioms, fixed expressions, and proverbs is generally connected with the culture and history of the people who speak that language as native speakers. Due to this fact, the expressions are not exactly the same in different languages. This creates a big challenge for  MT, since it is often not just literal or surface word translation and these kinds of translation do not convey the meaning in the target language. 
%MWEs are one of the common means of expressing idioms, fixed expressions (e.g. kick the bucket), customised expressions (e.g. once upon a time),  proverbs, and more. In light of this, it is  natural to think of the combination of MT and MWEs, e.g., how to correctly translate MWEs in  MT to make meaning-equivalent translations. This includes how to combine MWE identification and MWE prepossessing into MT models and systems to achieve better translation.

%There is a workshop on Multi-word Units in Machine Translation and Translation Technology (MUMTTT) which was held every two years, and the 4th edition was held in 2019\footnote{http://www.lexytrad.es/europhras2019/mumttt-2019-2/}.

%In our work we assume that bilingual and multi-lingual MWEs corpora can be helpful to many NLP tasks such as MWE detection, MT, cross-lingual question answering and information extraction, etc.

\section{Related Work in Corpus Construction}
In this section, we introduce some related work on MWE corpus construction  to advance  MWE research. This includes ``MWE aware English Dependency corpus" from  LDC\footnote{https://catalog.ldc.upenn.edu/LDC2017T01}, where they annotated  English compound words in the corpus as one kind of MWE to facilitate the constituency and dependency parsing task; and the annotation of English MWEs in web reviews data \cite{schneider-etal-2014-comprehensive} \footnote{http://www.cs.cmu.edu/~ark/LexSem/}, where they  hand-annotated online review data with comprehensive MWEs including English noun, verb, and preposition super-senses (tags include communication, group, stative, location, possession, etc.). However, both these  MWE corpus construction works are monolingual tasks and focus on English only.

There is some multilingual MWE corpus construction from the PARSEME research project in \cite{Savarytv2018vMWE}, which includes 18 European languages. However, the constructed corpus focuses on one kind of MWE (verbal MWE), is not parallel, and the size of the data varies very much from language to language (some languages have only hundreds of sentences). We built a multilingual MWE database consisting of parallel phrases that can be used to extended NLP tasks, such as translation, extend to non-European languages (e.g. Chinese), and enlarge the size into hundreds of thousands and millions of pairs.

To build our MultiMWE corpus, we used the MWE extraction pipeline from \cite{rikters2017mwe}\footnote{https://github.com/M4t1ss/MWE-Tools}. We extend the extraction work into language pairs such as German-English and Chinese-English to assess the impact of MWEs on the NMT task in general and contributed the corresponding MWE extraction patterns of tested languages. Furthermore, the extracted MWEs from our experiments are freely available for MT and NLP researchers to use for their own tasks. However, the extracted MWE candidates from this framework is only the \textit{continuous} type. In  follow up work, we will design some patterns or other models to extract  discontinuous MWEs, for instance, ``apple of \textit{someone's} eyes", ``pick \textit{someone} up", and ``take \textit{something} into account", etc.

\section{MultiMWE Extraction Process}

In this section, we present the MultiMWE corpora construction, including German-English and Chinese-English parallel MWEs extraction and give some detailed procedures. 

\subsection{German-English}
\begin{figure*}[!t]
\centering
\includegraphics*[height=2.2in]{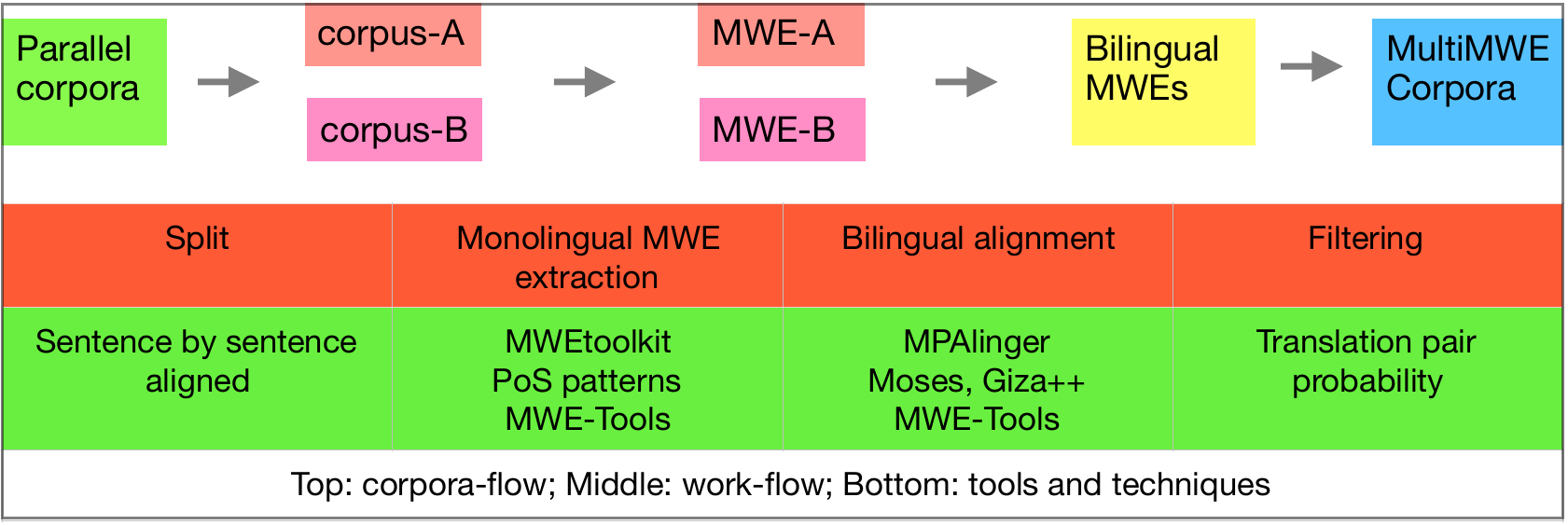}
%[height=3.1in,width=2.6in]
\caption{MultiMWE corpora extraction workflow.}
\label{fig:MultiMWE_workflow}
\end{figure*}

The root parallel corpus is from the WMT2017 German-English MT training task\footnote{http://data.statmt.org/wmt17/translation-task/preprocessed/}. This contains 5.8 million  German-English sentences. To create a suitable bilingual MWE corpus we adopted the following procedure (Figure \ref{fig:MultiMWE_workflow}).

\begin{itemize}
     \item Morphological tagging of De and En.
     \item Tagged De/En into XML format.
     \item Design MWE-patterns for De/En
     \item Extract Monolingual MWEs with MWEtoolkit
     \item Generate De-En lexicon translation probability files with Giza++ and Moses
     \item Align Bilingual MWEs with MPAligner
\end{itemize}

Firstly, Treetagger\footnote{https://www.cis.uni-muenchen.de/~schmid/tools/TreeTagger/} \cite{schmid1994treetagger} was used to tag English and German sentences with morphology information (PoS and lemmas). The English and German morphological tag-sets we used were from the BNC\footnote{\url{http://www.natcorp.ox.ac.uk/docs/c5spec.html}} and STTS\footnote{\url{https://www.cis.uni-muenchen.de/~schmid/tools/TreeTagger/data/stts_guide.pdf}} corpora. Secondly, we performed a mapping of the English POS patterns for MWE extraction from PENN used in \cite{rikters2017mwe} to BNC. We designed the German POS patterns for MWE extraction and then the English and German monolingual candidate MWEs were extracted using MWEtoolkit \cite{ramisch2015book_mwetoolkit} with the corresponding MWE patterns and the morphological corpus. 
Thirdly, the MWE-tools\footnote{https://github.com/M4t1ss/MWE-Tools} from \cite{rikters2017mwe} were used to convert the extracted two monolingual candidate MWE files into MPAligner format. Fourthly, we ran word alignment tools Giza++ and SMT platform Moses to get the lexical translation probability files of German-English both directions. The bilingual MWEs were aligned using MPAligner\footnote{\url{https://github.com/pmarcis/mp-aligner}} \cite{Pinnis2013mp_aligner} with the corresponding translation estimation probability. 
Finally, we examine the use of extracted bilingual MWEs in NMT experiments. We choose a state-of-the-art Transformer model for MT experiments, with the open package THUMT\footnote{http://thumt.thunlp.org}.

\begin{comment}

The procedure for MWE extraction and integration into NMT was as follows:

\begin{enumerate}
   \item Bilingual MWEs Extraction.
    \begin{enumerate}
     \item Morphological tagging of De and En.
     \item Tagged De/En into XML format.
     \item Design MWE-patterns for De/En
     \item Extract Monolingual MWEs with MWEtoolkit
     \item Generate De-En lexicon translation probability files with Giza++ and Moses
     \item Align Bilingual MWEs with MPAligner
    \end{enumerate}
   \item Aligned Bilingual MWE Candidates Pruning.
    \begin{enumerate}
     \item Extract Bilingual MWEs with threshold (using the MPAligner estimated MWEs alignment scores).
    \end{enumerate}
   \item Integrate MWEs into NMT
    \begin{enumerate}
     \item Concatenate MWEs into BPE encoded bilingual sources.   
     \item \textit{BPE-encode MWEs before Concatenate into BPE encoded bilingual sources}.   
     \item \textit{Update original vocabularies with MWEs for NMT}.
     \item THUMT package was applied for NMT trainer and translator.
    \end{enumerate}
\end{enumerate}

\end{comment}

\subsection{Chinese-English}

To the best of our knowledge, there is no openly available bilingual MWE corpus of Chinese-English, which means we needed to create our own again. For a root parallel training corpora, we use the publicly available WMT2018 Zh-En pre-processed (word segmented) data. However, due to computational limitations in the follow-up NMT experiment, we only use the first 5 million  parallel sentences as a training set.

The monolingual MWE extraction and bilingual MWE alignment procedure is mostly the same with German-English, except for some additional processing that we list as following.

\begin{itemize}
     \item PoS pattern design.
     \item Stop-word list preparation.
     \item Zh-En translation probability files.
\end{itemize}

For the PoS patterns that MWEtoolkit required for MWE extraction, we did a PoS pattern mapping from English to Chinese and added some other Chinese PoS tags that are apparently pointing to MWEs,  as shown in Table~\ref{tab:add_zh_mwe_patterns}. These include idioms, fixed expressions, personal names, place names and organization names. The Chinese tagset is from the Lancaster Corpus of Mandarin Chinese (LCMC)\footnote{https://www.lancaster.ac.uk/fass/projects/corpus/LCMC/}. 

In the bilingual MWE alignment step, MPaligner requires a stop word list for each language. We used some open-source packages to build a Chinese stop words file including the lists from Chinese leading IT company Baidu and NLP institutes in HIT University and Sichuan University. These packages are open source\footnote{https://github.com/stopwords-iso and https://github.com/goto456/stopwords}. We removed  duplicates when concatenating the word-lists together. There are 2,361 Chinese stop words in the merged file.

Again, we run Giza++ and Moses toolkit to get the Chinese-English lexicon translation probability files from both directions. These files will be used for bilingual MWE alignment by MPaligner.

\begin{table}
\begin{center}
\begin{tabular}{ l|l } \hline 
  i & idiom \\
  l & fixed expressions \\
  nr & personal name\\
  ns & place name\\
  nt & orgnization name\\ \hline 
\end{tabular}
\caption{Added Chinese Patterns for MWEs from the LCMC Tags}
\label{tab:add_zh_mwe_patterns}
\end{center}
\end{table}

\subsection{Bilingual MWE Filtering}

\begin{figure}[!t]
\centering
\includegraphics*[height=2.8in]{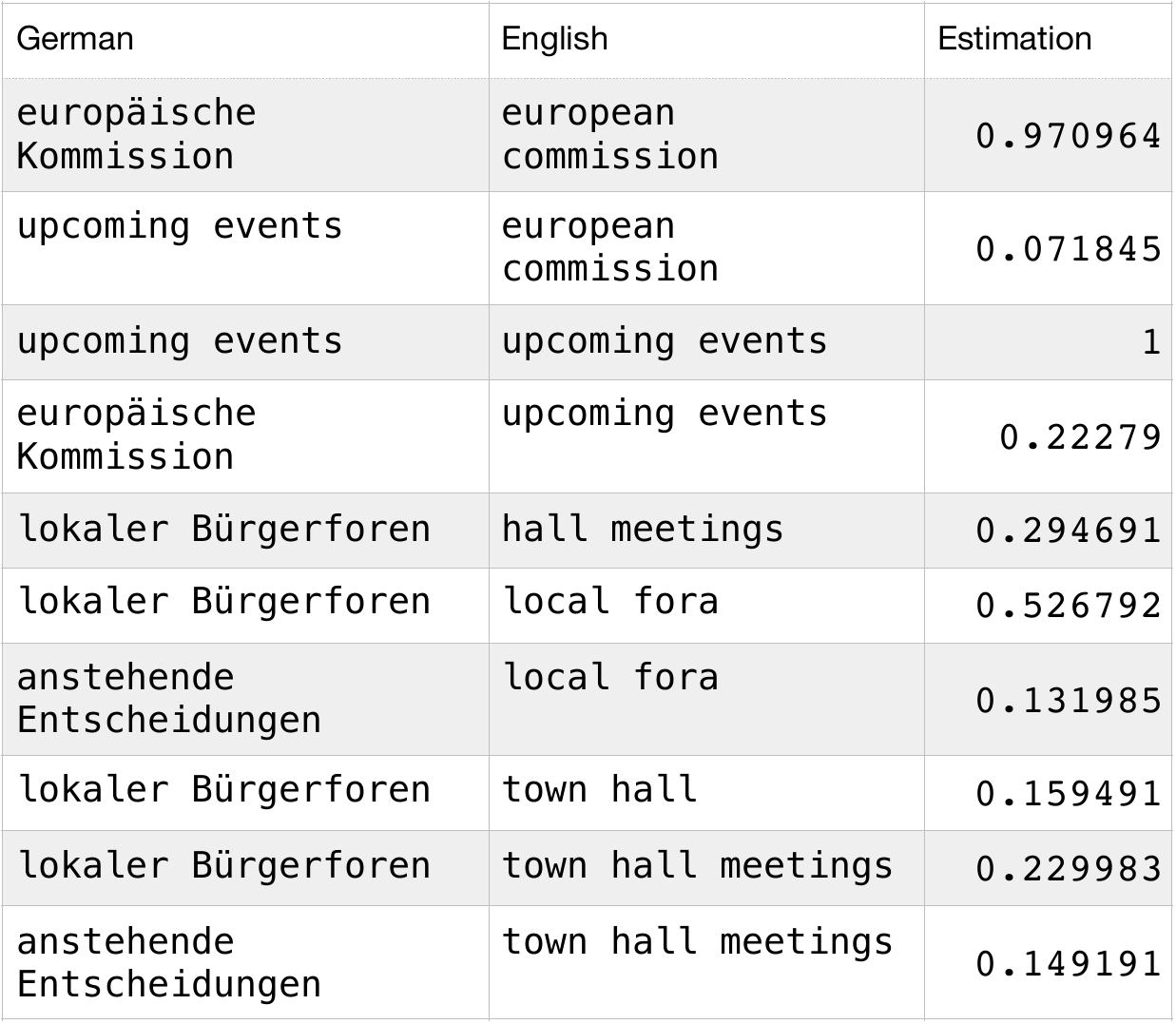}
%[height=3.1in,width=2.6in]
\caption{Samples of MPAligner aligned bilingual MWEs from 5 million English-German corpora. The higher the estimation scores, the better extracted bilingual candidates.}
\label{fig:MPaligner_sample_en_de}
\end{figure}

\begin{figure}[!t]
\centering
\includegraphics*[height=2.8in]{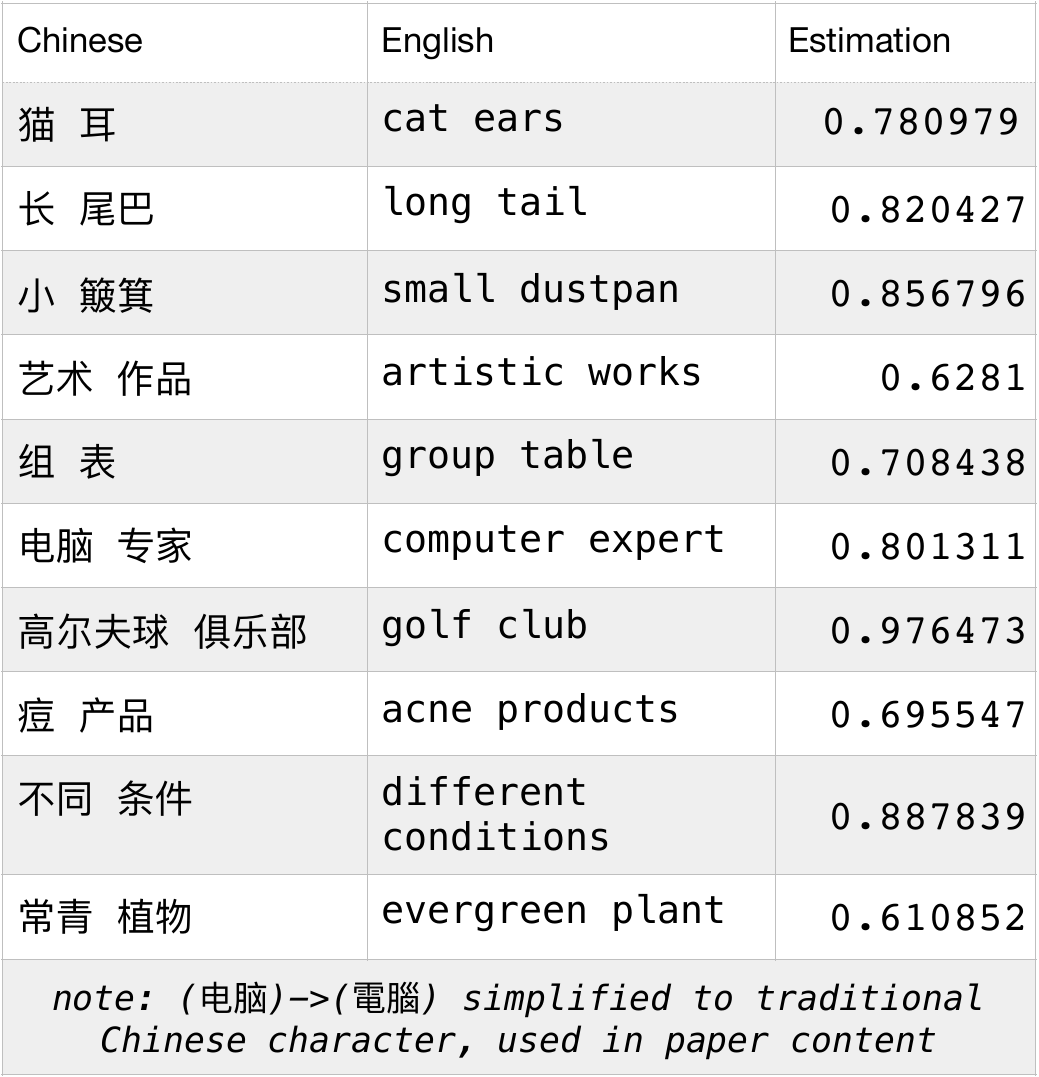}
%[height=3.1in,width=2.6in]
\caption{Extracted Zh-En MWEs without pruning. The extracted pair samples here are from the head of the file, which have good quality. }%The fifth column is the alignment estimation score, and the sixth column is the corresponding part-of-speech of the words from the Chinese side.
\label{fig:extracted_zh_en_mwe_sample}
\end{figure}

We manually examined the extracted bilingual MWEs and found that the MPAligner aligned bilingual MWEs have lots of noise, especially for German-English pairs. The output for German-English bilingual MWEs contains many candidates that have very low translation probabilities between 0 and 0.5. For example. The English term `European Commission' is aligned with German `Europäische Kommission' with 0.97 translation score, while `upcoming events' is also aligned to German `Europäische Kommission' with 0.22 translation probability which may be due to their co-occurrence and morphological patterns adj+noun (See Figure~\ref{fig:MPaligner_sample_en_de}). The extracted Chinese-English bilingual MWEs  generally have higher translation probability above 0.5 and are better quality (see Figure~\ref{fig:extracted_zh_en_mwe_sample}).

\begin{figure}[!t]
\centering
\includegraphics*[height=3.3in]{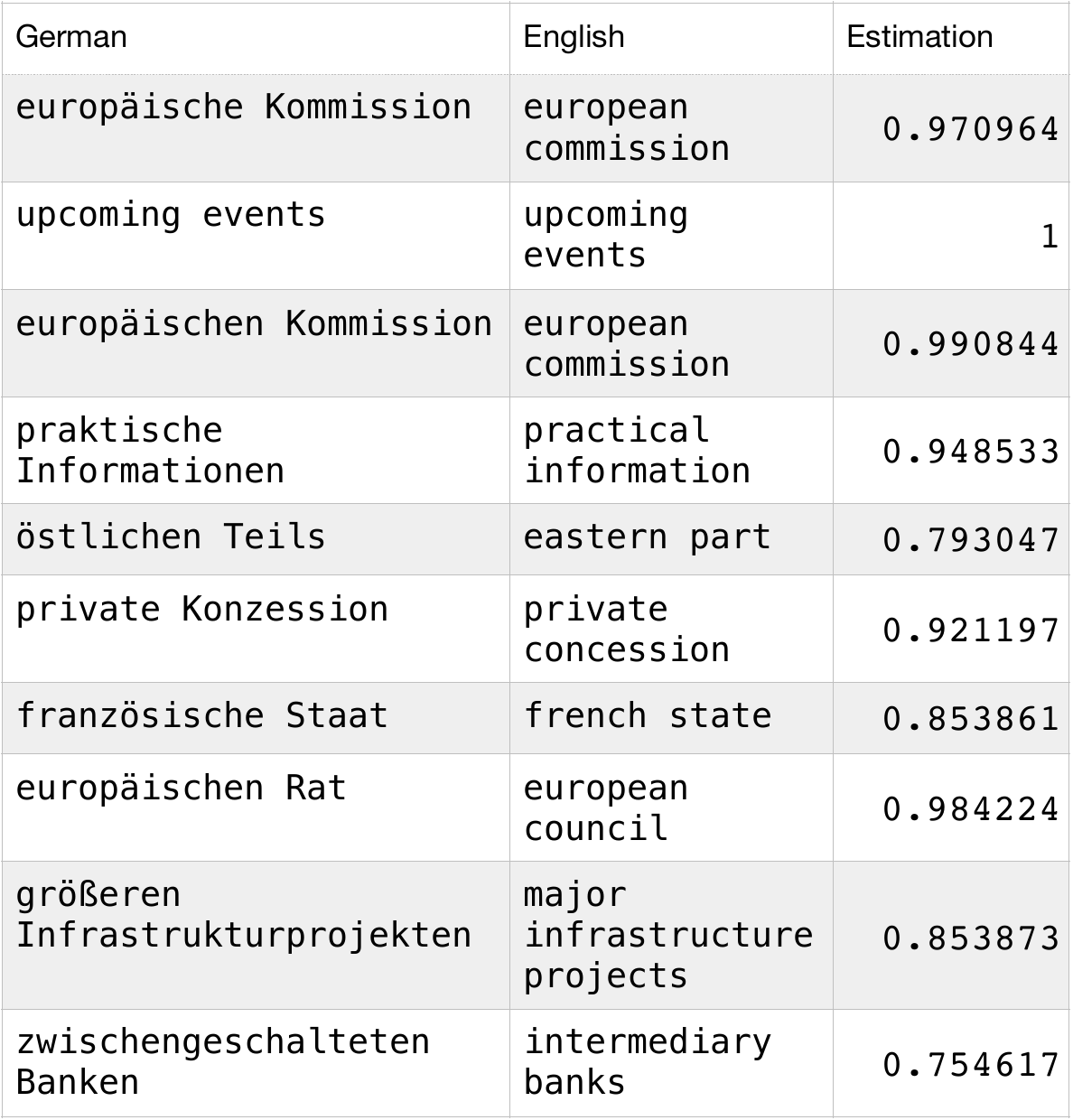}
%[height=3.1in,width=2.6in]
\caption{Samples of Bilingual MWEs after pruning with threshold 0.70. i.e.,  bilingual aligned MWEs with estimation score under 0.70 are removed.}
\label{fig:MPaligner_sample_en_de_pruning70}
\end{figure}

To filter out the low-quality bilingual MWE pairs, we chose two thresholds, i.e. 0.70 and 0.85, respectively in our experiments.
The initially extracted and subsequently aligned bilingual MWEs are 27,688,373 pairs and 172,900 pairs for German-English and Chinese-English respectively (see discussion section for this number differences). 
After pruning with alignment threshold 0.70 (see Figure \ref{fig:MPaligner_sample_en_de_pruning70} with samples) and 0.85, the German-English MWEs moved to 6,518,550 (23.5\% of original size) and 3,159,226 (11.4\%) pairs. The Chinese-English MWEs moved to 143,042 pairs (82.7\% of original size) with alignment threshold 0.85.
%6518550 ÷ 27688373 =0.23542553403
%3159226 ÷ 27688373 =0.11409937304
% 143042 ÷ 172900 =0.82731058415

\begin{CJK*}{UTF8}{bsmi}

From the examples in Figure \ref{fig:extracted_zh_en_mwe_sample} and Figure \ref{fig:MPaligner_sample_en_de_pruning70}, we can see that the extracted MWEs include some non-decomposable ones. For instance, the Chinese MWE ``簸箕" (Bòji) with two characters together means ``dustpan", and ``電腦" (Diànnǎo) \footnote{we use traditional Chinese characters overall in the paper content for consistency, also to solve the character encoding issues} means ``computer". However, if we split the two characters of any of them, it can not make the same meaning. The Chinese character ``電" means ``electricity", while ``腦" means ``brain". So the combined character sequence ``電腦" is a metaphor to describe ``computer". For decomposable MWEs that we extracted, there are ``european commission" and ``european council" as institutional names in Figure \ref{fig:MPaligner_sample_en_de_pruning70}.
%電腦
\end{CJK*}

\section{MWE+MT Experiments}
To verify the quality of our extracted bilingual MWEs, as one example, we apply them to NMT experiments as additional knowledge to influence  NMT learning. This is achieved by concatenating the extracted bilingual MWEs back to the original bilingual training corpus as additional ``translation pairs". We call the learning model with the extracted MWEs added to training corpus  `MWE+Base' and call the model with filtered MWEs ``MWEpruned(threshold)+Base", e.g. MWEpruned0.7+Base.

%To investigate the impact of MWEs on NMT for different language pairs, both for distant and for closer languages, and to estimate the potential impact of MWEs on NMT for untested language pairs, we first run the English-German and English-Chinese corpora as the initial language pair, and  future development will address more language pairs. 

%Similar to the work in \cite{rikters2017mwe}, for NMT learning model, we use the same bilingual corpus that we used to extract MWEs, which is 5.8 million sentences for German-English and 5 million sentences for Chinese-English.

The baseline NMT model is a state-of-the-art Transformer (THUMT-tensorflow) from \cite{thumt2017}. This implements the all-attention based NMT encoder-decoder structure developed by Google Brain \cite{google2017attention}. The sub-word unit translation BPE methodology \cite{SubwordNMT15Sennrich} is applied for the improvement of rare word translation. As a standard-setting, the BPE operations size is set to 32k for both German-English and Chinese-English training corpora; the vocabulary-threshold is set to 50, which means any word with frequency less than this threshold will be treated as an (out-of-vocabulary) OOV word; training set shuffling is applied by randomly relocating the order of each sentence; batch size is set at 6250. The encoder and decoder are set up with 7+7 layers.

\subsection{German-English MT}

The training corpus for NMT is the same as used for MWE extraction, 5.8 million parallel German-English sentences; the development and testing corpora are 3,003 and 2,169 parallel sentences respectively. 
To examine the external German-English MWEs that are available, we also set up one experiment where we added the 871 external MWE pairs into the training corpus. We call this ExterMWE871+Base.

% \caption{De-2-En NMT BLEU scores. ExterMWEtrain+Base means we added the 871 external de-en MWEs into the training set and learn the BPE and vocabulary from start.}

After the first 20k learning steps are applied, the evaluation scores are displayed in Table~\ref{tab:de2en_bleu_20k}. This result shows that even though in most n-gram matching the Baseline achieved better scores,  the overall BLEU score is lower than the MWE+NMT case. This is due to the Brevity-Penalty (BP) parameter and ratio factors, which means that the MWE+NMT model produced more reference like output than the Baseline model.   

It is strange that by adding 871 pairs of external De-En MWEs into the training set, the ExterMWE871+Base performance score is not higher than the baseline. The reasons could be: 1)  due to the added MWEs being too small in size compared with the 5.8 million training set. 2)  the external MWEs are kept as one (large) token instead of being split by the BPE model. 3) the external MWE pairs have many metaphor expressions, but such metaphors did not appear often in the training corpus, and can also mislead the learned model. 
%keep the external MWEs as no BPE. The original extracted de-en MWEs are 27,688,373 pairs, and the number decreased to 6,518,550 with the filter threshold 0.70. The number of de-en MWE pairs decreased to 3,159,226 with the threshold (0.85, 1.00) without including border score \footnote{The en-de MWE pair with alignment estimation score as 1.00 are actually both English phrases according to file head observation.}.  The bpeMWE+prun85to100+Base MWEs were BPE-encoded before adding them to training. One strange thing is that the BPE encoder applied to the extracted MWEs makes the MWE into full alphabet without word information at all. 
%To address this, we add the prun85to100 MWEs again without BPE encoding them.
% Another experiment seting ``ExterMWE871(NoBPE-dash)" means we even connect MWEs words with dask token to keep them together.

\begin{table*}
\begin{center}
\begin{tabular}{l|c|c|c|c|c|c|c} \hline 
   & \multicolumn{4}{c|}{n-gram scores} & \multicolumn{2}{c|}{Params}& Combine\\ \hline 
  models & 1-gram & 2-gram & 3-gram & 4-gram & BP & ratio & overall\\ \hline
  Baseline  & \textbf{63.3} & \textbf{35.2} & \textbf{21.4} & \textbf{13.5} & 0.942 & 0.944 &26.73 \\
  MWEpruned0.7+Base  &63.0 & 35.1 & 21.3 & \textbf{13.5}  & \textbf{0.952} &\textbf{0.953} & \textbf{26.87}\\
  ExterMWE871+Base &\textbf{63.3} &\textbf{35.2} & 21.2 &13.3  &0.929   &0.932 & 26.15 \\
 \hline \end{tabular}
\caption{De-2-En NMT BLEU Scores with 20k Transformer Learning Steps. }
\label{tab:de2en_bleu_20k}
\end{center}
\end{table*}

\subsection{Chinese-English MT}

For Chinese-English baseline NMT training, we also use the same corpora that were used for MWE extraction, 5 million parallel Zh-En sentences. The development (newsdev2017) and testing (newstest2017) corpus for NMT were from WMT2017, 2002 and 2001 parallel sentences respectively. %For BPE technology, we use \textit{32k BPE operations}\footnote{default setting from the THUMT package} that are learned from the bilingual training set. 
%The sample head content of the 32k BPE operations as displayed in Figure \ref{fig:zhen_zhRen_bpe32k_sample}.

%In the experiments, RD+Base (Radical Downside) means only adding radical into the training set and radical as a new line not mixing with the original zh sentence. RB (Radical Behind) means only adding radical into the training set and radicals follow the original zh character as in the same line. $RD-TR$+Base (RD and Test Radical) means the testing source zh sentence was replaced only by radical. rd2 means radical decomposition degree as 2nd layer. RB-TVT(radical behind, training-validation-test)+Base means the radical was added behind the original character for all the training/validation/test set of Chinese. 
In the evaluation score Table~\ref{tab:zh2en_wmt18_bleu}, model MWEpruned0.85+Base means we pruned the extracted Zh-En MWEs with threshold 0.85, then we used the original BPE operators to encode the pruned MWE pairs and concatenated it to the BPE encoded training set. We used the same vocabulary file from the baseline model. The result shows that the pruned MWE pairs enhanced the model learning by producing \textbf{improved 3-gram and 4-gram BLEU scores} and yielding an overall higher score. This automatic score means that the MWE enhanced model can generally improve the chunk translation, i.e., the MT output sentences include more chunk of 3-gram and 4-grams words that match the reference sentences. Most likely, they are improved MWE translations.

%The original extracted zh-en MWEs are 172,900 pairs and the number decreased to 143,042 after filtering with 0.85 threshold. 
%FullRadical(rxd3) means we replace the original Chinese word sequences in training corpus with full radical sequences with deeper decompose (level 3) which leads to the Chinese sequence as full strokes.
% fig: _example_zhRen_zhen_32kvocab_page

\begin{CJK*}{UTF8}{bsmi}

\begin{figure}[!t]
\centering
\includegraphics*[height=4.4in]{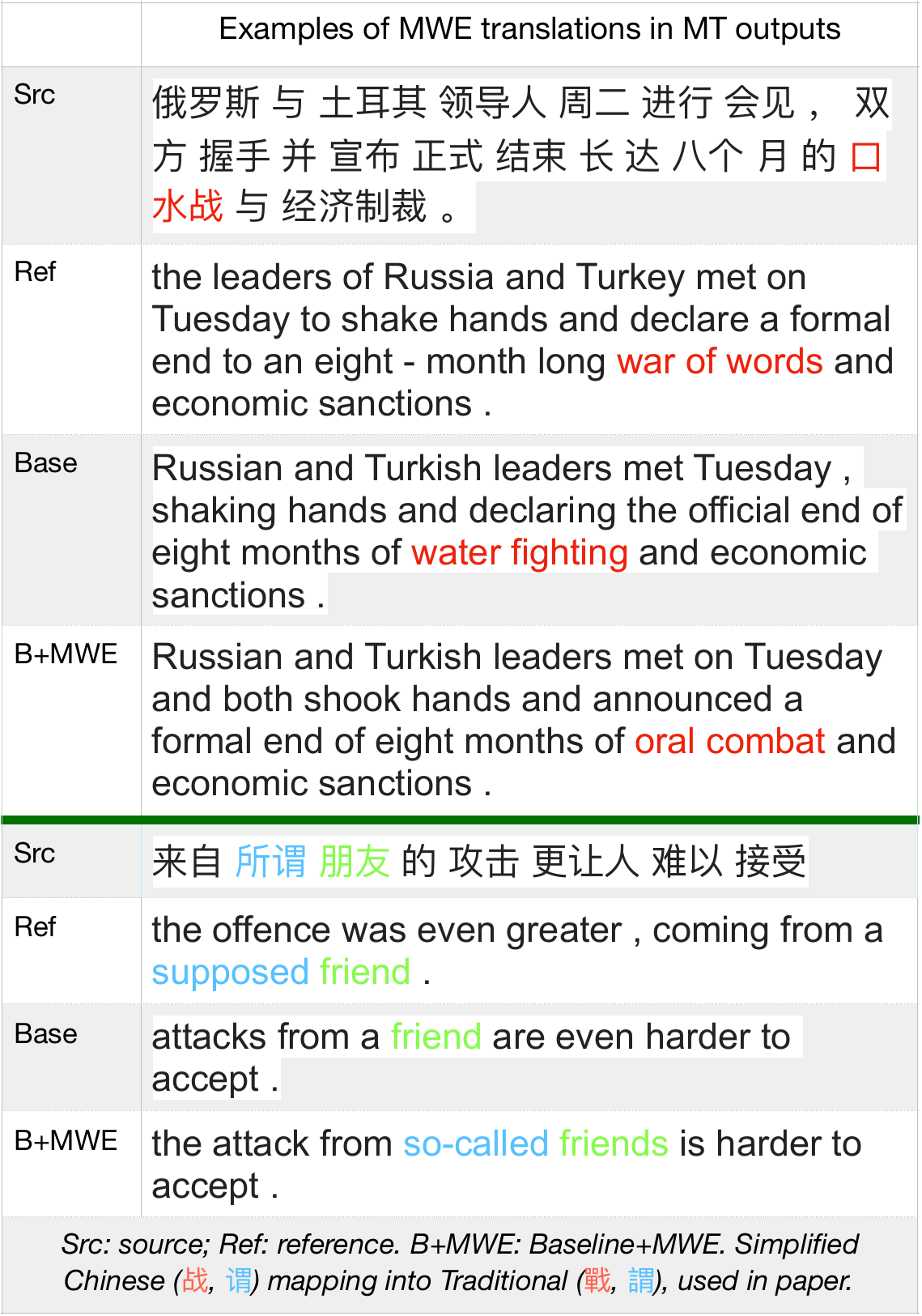}
%[height=3.1in,width=2.6in]
\caption{Examples of the translation outputs from Baseline model and model with filtered MWEs integrated into Baseline, with Chinese MWEs}
\label{fig:example_Base_vs_Basemwe85_zh2en}
\end{figure}

When we look inside the translation outputs from the baseline model and the MWE integrated model we found some Chinese MWEs that were not translated by Baseline and were translated properly by the MWEpruned0.85+Base model. Furthermore, some idiomatic MWEs that were translated literally by Baseline, were translated in a meaning preserved way by MWEpruned0.85+Base. See Figure~\ref{fig:example_Base_vs_Basemwe85_zh2en}, in the first example, Chinese ``口水戰" which means ``war of words" was translated into ``water fighting" by the Baseline, while it was translated into ``oral combat" in a proper way by MWE enhanced model. The Baseline translation is due to that this is a metaphor expression in Chinese using ``口水+戰" that is a combination of ``saliva" and ``war".

In the second example, ``所謂\, 朋友" which means ``supposed friend" is translated as ``friend" in Baseline model, and this lost the Chinese MWE ``所謂" which is used to express ``supposed" or ``so-called". The MT output yielded correct translation when we integrated the extracted bilingual MWEs back into the training corpus to enhance the learning.

However, both these two example sentences in Figure \ref{fig:example_Base_vs_Basemwe85_zh2en} showed that even though the MWE enhanced model produced better MWE translations, the BLEU scores of these two sentences do not improve correspondingly. The reason is that ``oral combat" can not match reference ``war of words" in the word surface form as used by BLEU metric; and ``so-called" can not match reference ``supposed" either.

\end{CJK*}

\begin{table*}
\begin{center}
\begin{tabular}{ l|c |c  |  c |c|c|c|c } \hline 
   & \multicolumn{4}{c|}{n-gram scores} & \multicolumn{2}{c|}{Params}&\\ \hline 
  models & 1-gram & 2-gram & 3-gram & 4-gram & BP&Ratio& overall\\ \hline
  Baseline  & \textbf{56.3} & \textbf{26.5} &14.3  & 8.2  & \textbf{0.9} &\textbf{0.905} &  18.39\\ \hline 
  MWE+Base & 55.9 &26.1 &14.3  &8.2  &0.884 &0.89 & 17.99    \\ \hline 
  MWEpruned0.85+Base & 55.9 &26.3 &\textbf{14.5}  &\textbf{8.4}  &0.899 &0.903 &\textbf{18.49}    \\ \hline 
 \end{tabular}
\caption{Zh-2-En NMT BLEU scores with 20k Transformer learning steps. }
\label{tab:zh2en_wmt18_bleu}
\end{center}
\end{table*}

%   (rd2)RD+Base & 53.6 &24.5 &13.0  &7.3  &0.950 &0.951 & 17.86   \\ \hline 
%  (rd2)$RD-TR$+Base & 53.3 & 24.3&13.1  &7.4  &\textbf{0.953} &\textbf{0.954} &17.96    \\ \hline 
%  (rbd1)RB-TVT+Base & 53.7 &23.6 &12.1  &6.7  &0.9 &0.905 & 16.08    \\ \hline 

\section{Discussions and Future Work}

In this work, we presented bilingual MWE corpora for German-English and Chinese-English, two typologically different languages, which we call MultiMWE-corpora. They cover 3,159,226 and 143,042 pairs of German-English and Chinese-English bilingual MWE entries after filtering. These corpora are freely available, and the size is much larger than the currently available bilingual MWE corpus. However, this current extraction procedure only generates continuous MWEs. In the future, we will design patterns to extract discontinuous MWEs or develop new extraction models.

In the current experiments, the German and Chinese PoS patterns for extracting MWEs are mapped from the English PoS tagset, via meaning equivalent alignment. In  future, we plan to design  German and Chinese patterns specifically for these languages and conduct some linguistic knowledge survey for this.

The NMT experiments for German-English and Chinese-English showed one example usage of the extracted bilingual MWEs, where they improved the automated translation evaluation scores slightly by BLEU metric in quantitative analysis, and assisted better MWE translations in qualitative analysis. By running the BLEU metric, the results are different from one language pair to another. In future work,
we will explore more automated metrics that can conduct better meaning equivalent evaluation such as LEPOR\cite{han2012lepor}, and further investigate the translation output in more detail, such as a human in the loop evaluations and looking at MWE translations in general.

For Chinese-English MWE for NMT, we will include Chinese radicals and strokes (decomposed from Chinese characters) \cite{HanKuang2018NMT} into the system, and investigate the performance with these linguistic features. % and change some model settings, such as extending the BPE operations into a larger number than 32k. 
%In the future experiments, we will include the combination of both radical and extracted MWEs together into zh-en NMT.

When we used MWEtoolkit for Chinese monolingual MWE candidate extraction there were some issues with the toolkit for this language, which meant we had to drop out some parts of the morphologically tagged corpus. This reduced the potential MWE numbers that can be produced by this procedure. In the future, we will look at this issue and fix the toolkit for the Chinese language. This will further extend our MultiMWE corpora size for the Chinese-English pair.

We make our extracted bilingual and multilingual MWEs corpora openly available. We believe that the MultiMWE corpora can be helpful for other multilingual NLP research tasks such as multi-lingual Information Extraction (IE), Question Answering (QA), and Information Retrieval (IR). For instance, those multi-lingual / cross-lingual tasks can take MultiMWE corpora as external dictionaries/knowledge into their models.

%Mapping from english, This will cause issues regarding  Chinese MWE patterns which shall be probably very different to English from the linguistic point of view. The further work of Chinese MWEs pattern design will be conducted with linguistic insight. 

In future work we will extend the MultiMWE corpora to other language pairs, including similar and distant languages, such as Russian-Japanese, English-French, etc. We will use the popular corpora Europarl\footnote{https://www.statmt.org/europarl/} for this purpose.

%cut: In \cite{parizi2018mwe}, the authors did some experimental analysis on whether the character level neural language models capture the knowledge of MWE compositionality in English and German languages, with two kinds of MWEs, i.e., noun compounds and verb-particle constructions. Though they did not apply this into NMT, it is very promising work to potentially indicate the possibility of conducting both \textbf{MWE detection and integration into MT with pure Neural models}. we plan to have this as part of our further experimental plans.

%\subsection*{Acknowledgements}
\section{Acknowledgements}
%\alan{Add adapt acknowledgement here}
The ADAPT Centre for Digital Content Technology is funded under the SFI Research Centres Programme (Grant 13/RC/2106) and is co-funded under the European Regional Development Fund. The input of Alan Smeaton is part-funded by Science Foundation Ireland under grant number SFI/12/RC/2289 (Insight Centre). We thank Matiss Rikters and Mārcis Pinnis for the supports of MWE-Tools and MPaligner, and the reviewers for valuable comments. LH thanks Paolo Bolzoni for helping experiments, Gültekin Cakir and Anna Weidmann for looking at German MWEs.
%LH thanks  Paolo Bolzoni, Matiss Rikters and M\={a}rcis Pinnis for help during  experiments, and thanks G\"{u}ltekin Cakir for checking some German MWEs.
% https://en.wikibooks.org/wiki/LaTeX/Special_Characters

\section{References}

% \nocite{*}
%\section{Bibliographical References}
%\label{main:ref}

\bibliographystyle{lrec}
\bibliography{lrec2020W-xample}

%\section{Language Resource References}
%\label{lr:ref}
%\bibliographystylelanguageresource{lrec}
%\bibliographylanguageresource{lrec2020W-xample}

\end{document}